\documentclass[conference]{IEEEtran}
\IEEEoverridecommandlockouts
\usepackage{cite}
\usepackage{amsmath,amssymb,amsfonts}
\usepackage[noend]{algpseudocode}

\usepackage{algorithm}
\usepackage{multirow}
\usepackage{graphicx}
\usepackage{textcomp}
\usepackage{xcolor}
\usepackage{subcaption}
\usepackage{makecell}
\usepackage{booktabs}
\usepackage{pdfpages}
\usepackage{hyperref}
\def\BibTeX{{\rm B\kern-.05em{\sc i\kern-.025em b}\kern-.08em
    T\kern-.1667em\lower.7ex\hbox{E}\kern-.125emX}}
\begin{document}

\title{Efficient Deep Learning for Biometrics: Overview, Challenges and Trends in Ear of Frugal AI
\thanks{}
}

\author{
\IEEEauthorblockN{1\textsuperscript{st} Karim Haroun}
\IEEEauthorblockA{\textit{LIASD Laboratory} \\
\textit{University of Paris 8}\\
Saint-Denis, France \\
\href{mailto:karim.haroun@univ-paris8.fr}{karim.haroun@univ-paris8.fr}}
\and
\IEEEauthorblockN{2\textsuperscript{nd} Aya Zitouni\textsuperscript{\dag}}
\IEEEauthorblockA{\textit{LIASD Laboratory} \\
\textit{National School
of Artificial Intelligence}\\
Algiers, Algeria \\
\href{mailto:aya.zitouni@ensia.edu.dz}{aya.zitouni@ensia.edu.dz}}
\and
\IEEEauthorblockN{3\textsuperscript{rd} Aicha Zenakhri\textsuperscript{\dag}}
\IEEEauthorblockA{\textit{LIASD Laboratory} \\
\textit{National School
of Artificial Intelligence}\\
Algiers, Algeria \\
\href{mailto:aicha.zenakhri@ensia.edu.dz}{aicha.zenakhri@ensia.edu.dz}}
\and
\IEEEauthorblockN{4\textsuperscript{th} Meriem Amel Guessoum}
\IEEEauthorblockA{\textit{LSI/USTHB} \\
\textit{National School
of Artificial Intelligence}\\
Algiers, Algeria \\
\href{mailto:meriem.guessoum@ensia.edu.dz}{meriem.guessoum@ensia.edu.dz}}
\and
\IEEEauthorblockN{5\textsuperscript{th} Larbi Boubchir}
\IEEEauthorblockA{\textit{LIASD Laboratory} \\
\textit{University of Paris 8}\\
Saint-Denis, France \\
\href{mailto:larbi.boubchir@univ-paris8.fr}{larbi.boubchir@univ-paris8.fr}}
\thanks{\textsuperscript{\dag}Aya Zitouni and Aicha Zenakhri contributed equally to this work.}
}

\maketitle

\begin{abstract}

\end{abstract}

\begin{IEEEkeywords}
 Recent advances in deep learning, whether on discriminative or generative tasks have been beneficial for various applications, among which security and defense. However, their increasing computational demands during training and deployment translates directly into high energy consumption. As a consequence, this induces a heavy carbon footprint which hinders their widespread use and scalability, but also a limitation when deployed on resource-constrained edge devices for real-time use. In this paper, we briefly survey efficient deep learning methods for biometric applications. Specifically, we tackle the challenges one might incur when training and deploying deep learning approaches, and provide a taxonomy of the various efficient deep learning families. Additionally, we discuss complementary metrics for evaluating the efficiency of these models such as memory, computation, latency, throughput, and advocate for universal and reproducible metrics for better comparison. Last, we give future research directions to consider.

\end{IEEEkeywords}

\section{Introduction}
\label{sec:introduction}

In this paper, we address the challenge of deploying resource-intensive deep learning models for biometrics in resource-constrained real-world security applications. We provide a structured overview and taxonomy of Efficient Deep Learning (EDL) techniques, surveying the state-of-the-art methods that aim to reconcile the high computational demands of advanced models with the practical needs of scalable, secure, and responsive biometric systems.

The widespread adoption of biometric technologies such as face, fingerprint and iris recognition has fundamentally transformed identity management and security protocols \cite{kilany2025comprehensive, adil2025securing}. Deep Learning (DL) is the cornerstone of this revolution, enabling super-human accuracy in tasks such as face verification on challenging benchmarks \cite{kilany2025comprehensive}. The scope of biometrics continues to expand, with ongoing research into a diverse set of physiological and behavioral modalities, each with distinct properties regarding universality, uniqueness, and robustness. This has made biometrics a reliable alternative to traditional authentication methods such as passwords and tokens \cite{al2025comprehensive}.

However, this remarkable performance comes at a significant cost. The pursuit of higher accuracy has led to increasingly complex, deep, and computationally expensive models, creating a substantial computational burden. These models are often characterized by high memory consumption, significant energy demands, and slow inference speeds \cite{liu2025meta}. This reality conflicts with the constraints of modern security applications, which require real-time processing on edge devices such as smartphones and surveillance cameras, where computational resources, battery life, and bandwidth are limited \cite{adil2025securing}. Furthermore, the centralization of processing in the cloud raises privacy concerns, as sensitive biometric data is transmitted over networks, and operational costs become prohibitive for large-scale deployment. Adding to this issue, complex models can be more vulnerable to adversarial attacks and their performance can degrade significantly when training data is insufficient.

\begin{figure*}[!t]
    \centering
    \includegraphics[width=0.99\linewidth]{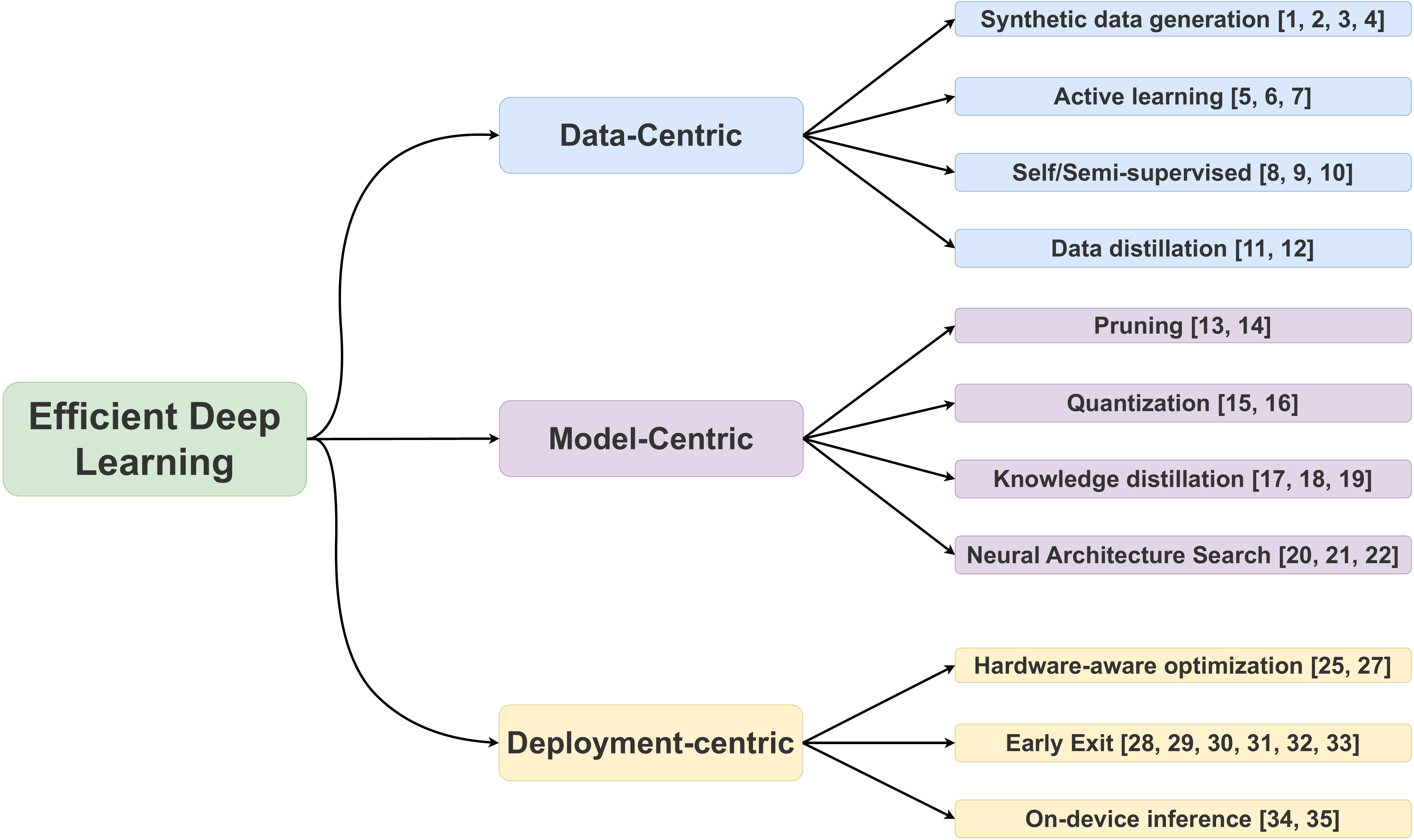}
    \caption{Overview taxonomy of Efficient Deep Learning (EDL)}
    \label{fig:placeholder}
\end{figure*}

This landscape requires a paradigm shift from pursuing higher accuracy to strategically optimizing the balance between performance and efficiency. We pose the central question: How can we maintain the high performance of deep learning for biometrics while drastically reducing its computational footprint? The answer lies in Efficient Deep Learning (EDL). EDL is not a compromise, but a necessary evolution for building the next generation of scalable, practical, and secure biometric systems. It represents a design philosophy focused on creating models and workflows that are resource-efficient in terms of computation, memory, and energy. As explored in this survey, EDL encompasses a range of techniques broadly categorized into Data-Efficiency, which reduces reliance on massive labeled datasets through methods like synthetic data generation and few-shot learning. Model-Efficiency, which creates smaller and faster models via compression, quantization, and efficient architecture design, and Compute/Energy-Efficiency, which optimizes deployment through hardware-aware optimization and on-device inference strategies. Recent work on Evidential Deep Learning (EDL) also offers promising pathways to achieve trustworthy predictions with limited data, a key concern in efficient learning. In this survey, we present many contributions that break down as follows:

\begin{itemize}
\item We organize a structured taxonomy and comprehensive survey of Efficient Deep Learning (EDL) techniques.

\item We systematically review the state-of-the-art in data, model, and computational efficiency, organizing a fragmented literature into a coherent analytical framework.

\item We contextualize these methods within biometric systems and discuss applications, open challenges, and promising research trajectories.

\end{itemize}

\begin{table*}[!t]
\caption{Summary of Efficient Deep Learning approaches}
\label{tab:edl_summary}
\centering
\renewcommand{\arraystretch}{1.3} 
\begin{tabular}{p{1cm} p{4cm} p{3cm} p{4cm} p{3cm}}
\toprule
\textbf{Category} & \textbf{Technique} & \textbf{Objective} & \textbf{Representative Methods} & \textbf{Limitations} \\
\midrule
\multirow{4}{*}{\makecell{\rotatebox{90}{\textbf{Data-Centric}}}} 
& Data Augmentation \& Synthetic Data & Increase dataset size \& diversity & Geometric transforms, Style transfer, GANs \cite{goyal2024systematic, paproki2024synthetic} & Quality of synthetic data may affect performance \\
& Active Learning & Reduce labeling cost & Uncertainty sampling, query-by-committee \cite{bayer2025activellm, li2024survey} & Requires iterative training cycles \\
& Self-/Semi-Supervised Learning & Leverage unlabeled data & Contrastive learning, Masked autoencoders \cite{chen2020simple, he2022masked} & Pretraining may be computationally expensive \\
& Data Distillation / Condensation & Compress dataset & Dataset distillation, coreset selection \cite{wang2018dataset, cazenavette2022dataset} & May not fully capture rare patterns \\
\midrule
\multirow{5}{*}{\makecell{\rotatebox{90}{\textbf{Model-Centric}}}} 
& Pruning & Remove redundant parameters & Unstructured/structured pruning \cite{han2015deep, li2020pruning} & Can reduce accuracy if aggressive \\
& Quantization & Lower numerical precision & 8-bit / low-bit weights \cite{jacob2018quantization, choukroun2020low} & May affect robustness, sensitive to adversarial attacks \\
& Knowledge Distillation & Train compact model from teacher & Student-teacher frameworks \cite{hinton2015distilling, gou2021knowledge} & Teacher must be high-quality; may propagate biases \\
& Efficient Architecture Design & Build inherently lightweight networks & MobileNets, SqueezeNets, EfficientNet, NAS \cite{tan2019efficientnet, zoph2018learning} & Limited capacity, may underperform on complex tasks \\
& Transformer Compression & Combine pruning, quantization, distillation & MobileViT, Pruned/Quantized ViTs \cite{tay2022efficient, chen2022mobilevit} & Trade-off between token reduction and performance \\
\midrule
\multirow{3}{*}{\makecell{\rotatebox{90}{\textbf{Deployment-Centric}}}} 
& Hardware-Aware Optimization & Adapt models to devices & Edge TPUs, microcontrollers \cite{han2016eie, millar2025energy} & Requires device-specific tuning \\
& Early-Exit / Dynamic Inference & Reduce inference for easy samples & BranchyNet, multi-exit networks \cite{teerapittayanon2016branchynet, li2023predictive} & Hard to optimize exit thresholds \\
& On-Device vs Cloud-Offloading & Balance computation \& latency & Edge inference, hybrid cloud-edge pipelines \cite{chen2019deepdecision, mao2017deep} & Dependency on network availability, heterogeneous devices \\
\bottomrule
\end{tabular}
\end{table*}

The remainder of this paper is organized as follows. Section \ref{sec:overview} offers an overview of Efficient Deep Learning and the key metrics for their evaluation. Section \ref{sec:sota} reviews the current state-of-the-art and Section \ref{sec:applications} provides some applications of EDL to biometrics. Section \ref{sec:discussion} discusses the overarching challenges and outlines promising future research directions. Finally, Section \ref{sec:conclusion} concludes the survey.

\section{Overview and background}
\label{sec:overview}

\subsection{The main approaches in EDL}

EDL models are designed to remain effective within limited resources. To translate this philosophy of frugality into practical strategies, it is useful to examine the principles that guide their design. First, data efficiency, second, model efficiency, and third, energy efficiency. Each of these addresses an aspect of resource usage, and together they provide a framework for developing EDL models that perform reliably even with constraints on data, memory, or computation.

\subsubsection{Data Efficiency}

Data efficiency focuses on minimizing the dependency of EDL on large datasets. EDL models seek to extract the maximum information from limited or imperfect data. Several techniques are applied to achieve this, and the choice of strategy depends on various parameters and the final usage of the data. The common  techniques in this category include data augmentation which  increase the numbers of training samples  by generation a variation of existing ones  given a modest sized dataset,  transfer learning which reuses knowledge from previously trained models to reduce data requirements for a new task, and semi-supervised and self-supervised learning methods allow models to learn from partially labeled or unlabeled datasets. Online  learning enables incremental learning from  new data that allows adaptation to environmental change. Finally, feature engineering and dimensionality reduction help identify and retain the most relevant information while minimizing computational complexity  

\subsubsection {Model Efficiency}
 
Model efficiency addresses the computational complexity and memory footprint of deep learning systems. The objective is to build models that are smaller, faster, and less demanding in computation while preserving accuracy. Each technique aims to optimize one or more evaluation metrics, such as latency, memory usage, and throughput. Common techniques include quantization, which represents model weights and activations with lower precision values to reduce memory footprint and faster computation. Pruning removes irrelevant neurons, weights, or entire structures such as layers, attention heads, or filters, thereby simplifying the model and reducing the number of operations required. Knowledge distillation trains a smaller and more compact model by using the output of a larger model as soft targets, allowing the student model to reproduce the behavior of the original system while using fewer resources.

\subsubsection{Energy Efficiency}

Energy efficient deep learning refers to the design of systems that minimize computational complexity and the environmental impact during training or inference. Various techniques optimize how and where deep learning models are executed, including specialized hardware, such as GPUs, TPUs, FPGAs, or ASICs, to reduce their energy consumption. By carefully aligning algorithms with hardware capabilities, these models can be more efficient while maintaining acceptable accuracy in a resource-constrained environment. 

\subsection{Evaluating the efficiency of models}

Evaluation metrics are essential to guide the development and evaluation of EDL models. They provide a complete way to assess how well a system meets frugality objectives by capturing multiple dimensions such as computational cost, throughput, latency, and energy consumption. Because no single metric can fully capture the behavior or trade offs of an EDL model, we rely on multiple complementary metrics to achieve a fair, comprehensive, and balanced evaluation. his multi-metric perspective ensures that gains in one area do not conceal losses in another.

\subsubsection{Computational complexity}

One way to evaluate the complexity of EDL models is to measure their computational complexity. This is quantified using FLOPs (Floating-Point Operations) or MACs (Multiply-Accumulate operations), which represent the number of basic arithmetic operations required for a forward pass of the model. Both metrics are related, since a MAC does two operations, when a FLOP is a single operation. This is summarized in the following formula :

\begin{equation}
\text{MACs} = 2 × \text{FLOPs}
\end{equation}

Finally, models with lower FLOPs or MACs demand less processing power, which not only speeds up inference, but also reduces the energy consumption.

\begin{figure*}[!t]
    \centering
    \includegraphics[width=0.9\linewidth]{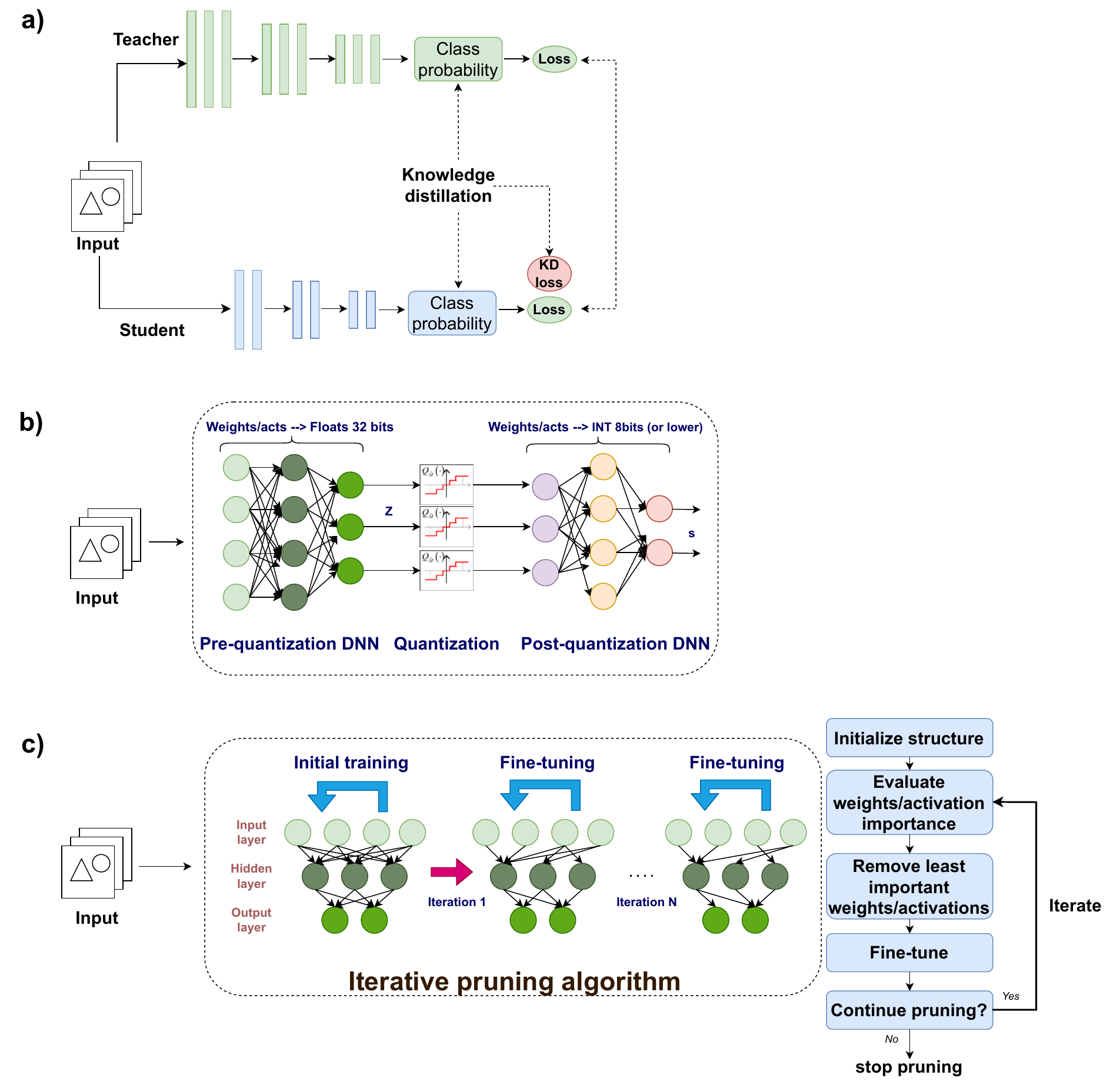}
    \caption{Overview diagram of three model-centric techniques. (a) Knowledge distillation, (b) quantization, (c) pruning.}
    \label{fig:compression_diagrams}
\end{figure*}

\subsubsection{Memory}

Memory is an important  metric for evaluating the efficiency of EDL models. The memory footprint of a model refers to the total space required to store its components during training or inference. This includes the model parameters, gradients, optimizer states, and intermediate activations, which can be expressed as follows:


\begin{equation}
    Total\_memory(bits) = (W + \nabla_{w}{\mathcal{L}} +S + A)\times b
\end{equation}

where $W$ denotes the parameters of the model, $\nabla_{w}{\mathcal{L}}$ the gradient of the loss function $\mathcal{L}$, $S$ the optimizer states, $A$ the activations of the model and $b$ the bit size of the model weights and activations. Note that using lower-precision representations for parameters and activations, such as 16-bit or 8-bit instead of 32-bit, reduces memory requirements for storage and the energy needed for data movement. Techniques like quantization leverage this principle to decrease memory usage without significantly affecting model accuracy.

\subsubsection{Latency}

Latency is the time required for a model to process a single input and produce an output, which makes it an important metric to evaluate the responsiveness of EDL models. Formally, we define latency as follows:

\begin{equation}
    \text{Latency} = \frac{\text{Total time}}{\text{Number of inferences }}
\end{equation}

\subsubsection{Throughput}

Throughput quantifies how many inferences an EDL model can make per unit of time, which measures the capacity of the model. It is defined as follows:

\begin{equation}
    \text{Throughput} = \frac{\text{Number of inferences}}{\text{Total time}}
\end{equation}

High throughput indicates efficient use of computational resources, which is important for efficient models to process more data per unit of time.

\subsubsection{Energy}
Energy efficiency measures how many computations or inferences a model can perform for a given amount of energy, this metric is important for devices with limited power. Energy efficiency can be expressed as the number of \emph{inferences per joule}, where high values indicate an energy-efficient system.
The energy consumed during inference depends on both the model \& hardware, it includes the energy needed to perform computations and the energy to move data 

\begin{equation}
 \text{Energy}_\text{total} = \text{Energy}_\text{data} + \text{Energy}_\text{MAC}
\end{equation}

The metrics used help evaluate the energy efficiency of an EDL model by providing an objective way to quantify its computational cost. By reducing computational complexity, we reduce the number of MAC operations, which directly decreases energy consumption, therefore offering a measurable assessment of the system’s power requirements.  

Defining appropriate metrics is fundamental to evaluate EDL models rigorously, as such metrics provide a quantitative and objective basis for assessing the degree of frugality achieved by a model. Equally important is the adoption of universal metrics that facilitate consistent comparisons between different algorithms based on standardized evaluation criteria. This practice ensures that advances in EDL remain cumulative, transparent, and methodologically stable in applications. Moreover, considering orthogonal metrics is necessary to prevent the optimization of a single dimension of frugality from inducing performance degradations in another. For example, a model may have a reduced computational complexity through pruning techniques, yet, this improvement can come at the expense of memory footprint if additional decision functions are added to the model.

\section{State-of-the-Art in Efficient Deep Learning}
\label{sec:sota}

There are three main efficient deep learning categories. First, data-centric techniques, second, model-centric techniques, and third, deployment-centric techniques. These approaches aim to reduce computational complexity, memory footprint, and annotation costs while minimizing the accuracy drop in models. Figure \ref{fig:placeholder} presents an overview of our taxonomy, while Table \ref{tab:edl_summary} provides an in-depth analysis of the methods surveyed.

\subsection{Data-Centric Techniques}

Data-centric techniques improve model efficiency by optimizing the data rather than modifying the network architecture. A widely adopted approach is synthetic data generation and augmentation, which creates artificial data to supplement or replace real-world datasets. Methods include geometric transformations, style transfer, and generative approaches such as GANs, enabling models to generalize effectively with limited labeled data \cite{goyal2024systematic, mumuni2024survey, paproki2024synthetic, shorten2019survey}.


\begin{algorithm}[h]
\caption{Data-Centric techniques}
\label{alg:data_centric}
\begin{algorithmic}[1]
\State \textbf{Input:} Initial model $\mathcal{M}$, labeled data $\mathcal{D}$ unlabeled pool $\mathcal{U}$, budget $B$
\While{$B>0$}
    \State $S \gets \operatorname*{argmax}_{x \in \mathcal{U}} \ \phi(x, \mathcal{M})$ \Comment{select data that max$(\phi)$}
    \State $\mathcal{L} \gets \operatorname{Label}(S)$
    \State $\mathcal{D} \gets \mathcal{D} \cup \{(S, L)\}$
    \State $\mathcal{M} \gets \operatorname{Train}(\mathcal{M}, \mathcal{D})$
    \State $B \gets B - \operatorname{Cost}(S)$
\EndWhile
\State \textbf{return} $M$
\end{algorithmic}
\end{algorithm}

Active learning allows the model to selectively query the most informative data points for labeling, which reduces annotation costs while preserving accuracy \cite{bayer2025activellm, li2024survey, maas2011learning}.

Self-supervised and semi-supervised learning leverages large amounts of unlabeled data to pre-train models or learn meaningful representations, minimizing the dependence on fully labeled datasets \cite{liu2021self,chen2020simple,he2022masked}.

Finally, data distillation or condensation techniques create small synthetic datasets that encapsulate the essential information of much larger datasets, enabling faster training and reduced storage requirements \cite{wang2018dataset, cazenavette2022dataset}. algorithm \ref{alg:data_centric} describes the general overview of these data-centric methods. Starting from an initial model $\mathcal{M}$ and labeled set $\mathcal{D}$, the method iteratively selects the sample $S \in \mathcal{U}$ that maximizes the acquisition score $\phi(x,\mathcal{M})$ under a labeling budget $B$. After querying its label and updating $\mathcal{D}$, the model $\mathcal{M}$ is retrained, and the process repeats until $B = 0$.

\subsection{Model-Centric Techniques}

Model-centric techniques directly reduce the computational complexity and memory footprint of deep networks. Model compression includes several strategies. Knowledge distillation, as shown in Figure \ref{fig:compression_diagrams}.a, transfers knowledge from a large teacher model \(T\) to a compact student model \(S\) by training \(S\) to mimic the output distribution of \(T\) through a distillation loss function \(\mathcal{L}_{\text{KD}}\) \cite{hinton2015distilling,gou2021knowledge,cheng2022model}.

Quantization, as shown in Figure \ref{fig:compression_diagrams}.b, reduces the full numerical precision of the parameters \(\mathbf{W}\) or the activations \(\mathbf{A}\), to a lower-bit representation \(\mathbf{W}_q\) or \(\mathbf{A}_q\), respectively. This is done through a quantization function \(Q(\cdot)\) that constrains values to a discrete set defined by a bit-width \(b\) \cite{jacob2018quantization,choukroun2020low}.

Pruning techniques, as shown in Figure \ref{fig:compression_diagrams}.c, iteratively remove redundant parameters or activations from a model during training by applying a binary mask \(\mathbf{M}\) to the weight tensor \(\mathbf{W}\) or the activation tensor \(\mathbf{A}\), where \(\mathbf{M}\) is derived from a saliency criterion \(\mathcal{C}\) \cite{han2015deep, li2020pruning}.

\begin{algorithm}[h]
\caption{Model-Centric techniques}
\label{alg:model_centric}
\begin{algorithmic}[1]
\State \textbf{Input:} initial model $\mathcal{M}$, dataset $\mathcal{D}$, efficiency budget $B$
\While{$\operatorname{Cost}(\mathcal{M}) > B$}
    \State $\mathcal{M}' \gets \operatorname{Adapt}(\mathcal{M})$ \Comment{modify structure/parameters}
    \State $\mathcal{M}' \gets \operatorname{Train}(\mathcal{M}', \mathcal{D})$
    \If{$\mathcal{A}(\mathcal{M}') \ge \mathcal{A}(\mathcal{M})$ \textbf{ and } $\operatorname{Cost}(\mathcal{M}')<\operatorname{Cost}(\mathcal{M})$}
        \State $\mathcal{M} \gets \mathcal{M}'$
    \EndIf
\EndWhile
\State \textbf{return} $\mathcal{M}$
\end{algorithmic}
\end{algorithm}

Efficient neural architecture design focuses on the construction of inherently lightweight networks such as MobileNets or SqueezeNets, or using Neural Architecture Search (NAS) to automatically discover resource-efficient architectures \cite{tan2019efficientnet,zoph2018learning,liu2019darts}. In transformer models, pruning, quantization, and distillation can be combined to produce efficient variants \cite{tay2022efficient,chen2022mobilevit}. Figure \ref{fig:compression_diagrams} provides an overview of three main model-centric techniques, namely knowledge distillation, quantization and pruning. Additionally, we describe a pseudo-code that encompasses these methods in algorithm \ref{alg:model_centric}. Starting from an initial model $\mathcal{M}$ and a dataset $\mathcal{D}$, the method iteratively adapts the model to obtain $\mathcal{M}'$ and retrains it while $\operatorname{Cost}(\mathcal{M}) > B$, progressively improving efficiency until the budget constraint is met.

\subsection{Deployment-Centric Techniques}

Deployment-centric techniques optimize models for practical inference constraints. Hardware-aware optimization adapts models to specific devices such as edge TPUs or microcontrollers, reducing latency and energy consumption \cite{han2016eie, millar2025energy, bouzidi2023hadas}.

Early-exit networks introduce intermediate classifiers, allowing simple samples to exit early, reducing inference time without a substantial accuracy loss \cite{teerapittayanon2016branchynet,huang2020multi, addad2025balancing, li2023predictive, laskaridis2020hapi, rognlien2022hardware}.

Finally, strategies for on-device inference versus cloud-offloading balance computation between local devices and remote servers by considering latency, energy, and bandwidth constraints \cite{chen2019deepdecision,mao2017deep}. We propose a unified pseudo-code in algorithm \ref{alg:deployment_centric} for deployment-centric techniques. Given a trained model $\mathcal{M}$ and a hardware profile $H$, the method selects the runtime policy $\pi^\star \in \Pi$ that maximizes the model performance $\mathcal{A}(\mathcal{M},\pi)$ while satisfying the latency constraint $\tau(\mathcal{M},\pi,H) \le L_{\max}$. The model is then deployed on $H$ using the selected policy $\pi^\star$.

\section{Applications of EDL in Biometrics}
\label{sec:applications}


\subsection{On-Device Biometric Authentication}

Running face, fingerprint, or behavioral biometrics directly on consumer devices instead of external servers improves privacy, reduces latency, and enables offline authentication. Lightweight deep metric learning deployed on edge hardware has demonstrated practical mobile authentication capabilities, delivering high identification accuracy while operating within restricted compute budgets \cite{wang2020framework,kokal2023deep}.

\subsection{Real-Time Video Surveillance on Edge Devices}

Dynamic situations such as person identification, re-identification, and anomaly detection often rely on video streams from resource-limited embedded cameras. Edge-based learning pipelines now support real-time biometric monitoring using reduced-complexity neural models, minimizing both bandwidth use and cloud-processing costs, while preserving temporal consistency for security monitoring. \cite{heo2024resource}

\subsection{Liveness Detection for Spoof Prevention}

Efficient anti-spoofing systems are needed to prevent presentation attacks involving printed photos, videos, masks, or replayed digital content. Recent lightweight liveness detection architectures achieve robust performance in mobile and embedded environments by combining compact feature extractors with efficient temporal cues such as micro-motions and depth estimations \cite{shinde2025enhanced, sharma2023survey}.

\begin{algorithm}[!t]
\caption{Deployment-Centric techniques}
\label{alg:deployment_centric}
\begin{algorithmic}[1]
\State \textbf{Input:} trained model $\mathcal{M}$, hardware profile $H$, latency constraint $L_{\max}$
\State $\Pi \gets \{\pi_1, \pi_2, \dots\}$ \Comment{candidate runtime policies}
\State $\pi^\star \gets \operatorname*{argmax}_{\pi \in \Pi} \ \mathcal{A}(\mathcal{M},\pi)$ \ \text{s.t.} \ $\tau(\mathcal{M},\pi,H) \le L_{\max}$
\State \textbf{deploy} $(\mathcal{M}, \pi^\star, H)$
\State \textbf{return} $\pi^\star$
\end{algorithmic}
\end{algorithm}

\subsection{Biometrics in Resource-Constrained Environments}

Deployments in remote areas, smart locks, or industrial IoT devices require biometric systems that function reliably with limited connectivity and low-cost hardware. Multiple studies show that compact biometric inference models and adaptive multimodal fusion can provide operational security when computing, memory, and power budgets are minimal \cite{yang2021biometrics,zeeshan2025continuous}.

\subsection{Privacy-Preserving Biometric Templates}

Reducing the computational footprint of secure biometric pipelines also affects privacy. Lightweight feature extractors enable the generation of compact, irreversible biometric templates that minimize storage overhead while resisting inversion and cross-matching attacks. These approaches align with modern privacy regulations and support authentication in decentralized environments \cite{prakasha2025privacy,abdullahi2024biometric}.


\section{Discussion and Research Directions}
\label{sec:discussion}

\subsection{Synthesis of EDL approaches}

Our survey of the state-of-the-art reveals several important trends. First, data-centric methods such as synthetic data generation, self-supervised learning, and active learning are important to overcome the scarcity of labeled biometric datasets, particularly for rare traits or new modalities \cite{goyal2024systematic, chen2020simple, li2024survey}. Second, model-centric techniques, including pruning, quantization, knowledge distillation, and lightweight architecture design, are highly effective in deploying accurate models on resource-constrained devices \cite{hinton2015distilling, han2015deep, tan2019efficientnet}. Third, deployment-centric strategies, such as early-exit networks, hardware-aware optimization, and adaptive on-device versus cloud inference, provide practical pathways to minimize latency, energy consumption, and bandwidth requirements \cite{teerapittayanon2016branchynet, millar2025energy, chen2019deepdecision}.

This indicates that combining data augmentation, model compression, and deployment-aware strategies is most promising for achieving efficiency when applied to biometric systems. For instance, knowledge distillation combined with quantization enables lightweight face recognition models suitable for real-time, on-device authentication, while synthetic data generation and active learning help alleviate limitations imposed by scarce labeled samples.

\subsection{Challenges and open problems}

Despite significant progress in recent years, there are still many challenges that persist in the implementation of efficient deep learning for biometric systems.

\subsubsection{The Accuracy-Efficiency trade-off} 

The Pareto frontier between model accuracy and resource consumption remains a challenge. Although techniques like quantization and pruning achieve substantial compression, they often lose precision in challenging biometric samples, such as low-quality fingerprints or occluded faces. The development of lossless compression methods or techniques that minimize accuracy degradation under extreme resource constraints represents an open research problem.

\subsubsection{Generalization vs. Specialization}

Efficient models optimized for deployment often show reduced generalization capability. Indeed, models pruned or quantized for one hardware platform may perform poorly on others, and those trained on constrained datasets struggle with demographic diversity or environmental variations. This specialization-generalization tradeoff is particularly present in biometrics, where systems must maintain performance across diverse populations.

\subsubsection{Security of EDL models} 

Lightweight models may introduce unexpected security vulnerabilities, such as in quantization, where they reduce the resilience of models against adversarial attacks by limiting numerical precision. Other techniques, such as knowledge distillation, can transfer teacher model vulnerabilities to student networks. The security implications of efficient techniques for biometric template protection and model inversion attacks require thorough investigation.

\subsubsection{Standardization and benchmarking} 

The field suffers from inconsistent evaluation protocols, with disparate datasets, hardware platforms, and efficiency metrics that hinder fair comparison. The absence of standardized benchmarks for efficient biometric systems complicates performance assessment and slows their adoption.

\subsubsection{Hardware-Software co-design} 

Most current approaches optimize algorithms independently of target hardware. However, efficiency gains require co-design of algorithms and hardware architectures, particularly for emerging platforms like neuromorphic processors and in-memory computing systems.

\subsection{Promising research directions}


\subsubsection{Dynamic and adaptive inference} 

One promising direction is the development of models that adjust their computational graph based on input hardness and system resources. For instance, Early-exit networks represent an initial step, but more advanced approaches could dynamically select model sub-networks, precision levels, or even architectural variants in real-time.

\subsubsection{Neuro-Symbolic AI for efficiency} 

Recently, we have witnessed the emergence of hybrid architectures that combine neural networks with symbolic reasoning. These architectures offer the potential for more data-efficient learning and improved interpretability. Although they are still applied in small tasks, applying them to biometric related problems could offer a promising direction.

\subsubsection{Automated efficient design} 

An interesting category we have not discussed in this short survey is Neural Architecture Search (NAS), which is a search-based compression method where the goal is to automatically search for optimized efficient architectures with minimal human intervention. NAS could be more involved in biometric applications, as it offers a heuristic multi-objective approach that balances accuracy, latency, and energy consumption for a given application.

\subsubsection{Explainability and fairness} 

As models become more efficient, it is important to ensure that they do not amplify biases or become opaque black boxes. Research should focus on developing efficient, interpretable models and fairness-aware compression techniques that maintain equitable performance across demographics.



Efficient deep learning represents not just an optimization tool, but a paradigm shift for the sustainable advancement of biometric security systems. As biometric applications proliferate across edge devices, IoT systems, and resource-constrained environments, the development of models that balance performance with resource constraints becomes imperative. Integration of efficiency considerations when designing the model, from data collection and architecture design to deployment and updates, will define the next generation of privacy-preserving, scalable, and robust biometric systems. The research community must continue to bridge the gap between computational theory and practical deployment, ensuring that efficiency advancements translate to improved security, accessibility, and reliability in real-world biometric applications.

\section{Conclusion}
\label{sec:conclusion}

In this paper, we provided an overview of Efficient Deep Learning (EDL) and the challenges it faces in the context of biometric systems and security applications. We classified EDL techniques into three main categories, namely, data-centric, model-centric, and deployment-centric approaches, and discussed their respective strategies, including data augmentation, knowledge distillation, model compression, and hardware-aware optimization. We highlighted the trade-offs between accuracy, computational efficiency, and deployment constraints. Furthermore, we identified open challenges such as generalization to unseen environments, robustness against adversarial attacks, and standardization of benchmarks.

By synthesizing current methods and challenges, this short survey aims to provide the biometrics and security community with a structured introduction to understand, evaluate, and adopt EDL techniques. We hope it will inspire future research to design lightweight, secure, and energy-efficient deep learning models that can be implemented effectively in practical biometric applications.

\bibliographystyle{IEEEtran}
\bibliography{references}

\end{document}